\documentclass[10pt,twocolumn,letterpaper]{article}
\usepackage[pagenumbers]{wacv}

\usepackage{times}
\usepackage{epsfig}
\usepackage{graphicx}
\usepackage{amsmath}
\usepackage{amssymb}
\usepackage{booktabs}

\DeclareMathOperator*{\argmin}{arg\,min}
\usepackage[dvipsnames]{xcolor}
\usepackage{subcaption}
\usepackage{caption}
\newcommand{\ignore}[1]{}
\usepackage{algorithm}
\usepackage{algorithmic}
\usepackage[utf8]{inputenc}
\usepackage{stackengine}
\usepackage{bm}

\def\wacvPaperID{795} 
\def\confName{WACV}
\def\confYear{2025}

\usepackage{comment}

\usepackage[breaklinks=true,bookmarks=false]{hyperref}

\begin{document}

\title{LaFA: Latent Feature Attacks on Non-negative Matrix Factorization}

\author{
Minh Vu\\
Theoretical Division, LANL\\
Los Alamos, U.S\\
{\tt\small mvu@lanl.gov} \\
\and
Ben Nebgen\\
Theoretical Division, LANL\\
Los Alamos, U.S\\
{\tt\small bnebgen@lanl.gov} \\
\and
Erik Skau \\
Computational Sciences\\
{\tt\small ewskau@lanl.gov}
\and
Geigh Zollicoffer \\
Theoretical Division, LANL\\
Los Alamos, U.S\\
{\tt\small gzollicoffer@lanl.gov} \\
\and
Juan Castorena\\
Computational Sciences, LANL\\
Los Alamos, U.S\\
{\tt\small jcastorena@lanl.gov} \\
\and
Kim Rasmussen \\
Theoretical Division, LANL\\
Los Alamos, U.S\\
{\tt\small kor@lanl.gov}
\and
Boian S. Alexandrov \\
Theoretical Division, LANL\\
Los Alamos, U.S\\
{\tt\small boian@lanl.gov}
\and
Manish Bhattarai \\
Theoretical Division, LANL\\
Los Alamos, U.S\\
{\tt\small ceodspspectrum@lanl.gov} 
}




\maketitle
\thispagestyle{empty}

\begin{abstract}

As Machine Learning (ML) applications rapidly grow, concerns about adversarial attacks compromising their reliability have gained significant attention. One unsupervised ML method known for its resilience to such attacks is Non-negative Matrix Factorization (NMF), an algorithm that decomposes input data into lower-dimensional latent features. However, the introduction of powerful computational tools such as Pytorch enables the computation of gradients of the latent features with respect to the original data, raising concerns about NMF's reliability. Interestingly, naively deriving the adversarial loss for NMF as in the case of ML would result in the reconstruction loss, which can be shown theoretically to be an ineffective attacking objective. In this work, we introduce a novel class of attacks in NMF termed \textit{Latent Feature Attacks  (LaFA)}, which aim to manipulate the latent features produced by the NMF process. Our method utilizes the \textit{Feature Error} (FE) loss directly on the latent features. By employing FE loss, we generate perturbations in the original data that significantly affect the extracted latent features, revealing vulnerabilities akin to those found in other ML techniques. To handle large peak-memory overhead from gradient back-propagation in FE attacks, we develop a method based on implicit differentiation which enables their scaling to larger datasets. We validate NMF vulnerabilities and FE attacks effectiveness through extensive experiments on synthetic and real-world data.

\end{abstract}

\section{Introduction} 

Non-negative matrix factorization (NMF)~\cite{lee2001algorithms} is a versatile tool for multi-way data reconstruction through factorizing a matrix or a multidimensional array in a least-squares approach. As real-world data often exhibit \textit{multiple ways}, e.g., conditions, channels, spaces, times, and frequencies, the NMF can be an effective tool to extract salient features of those data. As such, NMF has become widely adopted across scientific fields such as psychology, chemistry, signal processing, computer vision, and bioinformatics~\cite{wang2012nonnegative}. 



However, as we will show, NMF's decomposed factors are susceptible to input's perturbations, known as adversarial noise, which are intentionally designed to disrupt feature extraction. These disruptions, termed adversarial attacks, can compromise the reliability of the salient feature extraction process. Standard NMF algorithms typically assume that data are sampled from a distribution with a low-rank model and zero-mean i.i.d. Gaussian noise. In many real-world scenarios, this assumption may not hold due to the presence of malicious attacks or anomalies, rendering the feature extraction process vulnerable to such noise.

The main contributions of this paper are: 
\begin{itemize}
    \item We examine the robustness of NMF against adversarial attacks and find it vulnerable despite its resilience in data reconstruction. Specifically, we introduce the \textit{Feature Error} loss to directly assess latent features generated by NMF. Through back-propagation attacks using the FE loss, we show that injecting small amounts of adversarial noise into the data can lead to significant distortions in the resulting latent features. The finding established a class of new attacks, called Latent Feature Attacks (LaFA) on NMF.

    \item Back-propagating FE attacks require backtracking NMF's iterative updates, demanding substantial peak-memory. To address this, 
    we utilize implicit differentiation to determine a direct expression for the gradients required for FE attacks, bypassing the need to back-propagate through NMF iterations. This approach reduced the peak-memory requirements, and removes the history dependence of the FE attack gradients.
    \item We confirm the susceptibility of NMF to feature attacks and illustrate the efficacy of our approaches through comprehensive experiments conducted on one synthetic dataset and four different real-world datasets: WTSI~\cite{huang2018msignaturedb} , Face~\cite{lee1999learning}, Swimmer~\cite{donoho2003does}, and MNIST~\cite{lecun2010}.
\end{itemize}

This manuscript is organized as follows. Sect.~\ref{sect:related} and \ref{sect:prelims} provide the related work and preliminaries. Sect.~\ref{sect:method} describes our proposed FE loss and the corresponding LaFA targeting extracted features. Our methods and technical claims are illustrated via synthetic experiments in Sect.~\ref{sect:example}. Sect.~\ref{sect:realworld} provides our experimental results on real-world datasets, and Sect.~\ref{sect:conclusion} concludes this paper.

\section{Related Work}\label{sect:related}
Adversarial attacks in ML have predominantly targeted supervised learning models with numerous studies demonstrating the susceptibility of these models to subtle, maliciously crafted perturbations~\cite{madry2017towards, goodfellow2014explaining}. However, the exploration of adversarial attacks in unsupervised settings, particularly involving techniques like NMF, has started to garner attention only in recent years. 

Recent research has increasingly focused on integrating adversarial learning with NMF, revealing both vulnerabilities and opportunities for enhancing robustness. 
In \cite{luo2020adversarial}, the authors introduce adversarial perturbations during the factorization process, uncovering potential manipulations and inherent weaknesses in traditional NMF algorithms. 
To address this, \cite{cai2021adversarially} proposes a training regime that incorporates adversarial examples to foster NMF models that maintain precise factorizations under adversarial conditions. 
Extending beyond the direct applications to NMF, \cite{chen2022towards} explores the effects of adversarial attacks on community detection algorithms, often rooted in matrix factorization principles. The findings illustrate that robust algorithmic strategies can mitigate even extreme adversarial attacks, suggesting pathways to more resilient community detection methods. Additionally, \cite{seyedi2023elastic} merges deep learning with NMF to tackle matrix completion tasks, incorporating elastic adversarial strategies to assess and improve the robustness of learned patterns against deliberate noise. 

While the aforementioned studies primarily focus on defensive schemes within the context of NMF, especially during training, our work diverges by critically examining the robustness of NMF under adversarial attacks. We introduce novel techniques specifically designed to compromise feature integrity, marking a pioneering effort in executing targeted attacks within the realm of unsupervised learning. 

\section{Preliminaries} \label{sect:prelims}




NMF is particularly notable for its application in data with inherent non-negativity, where it decomposes a non-negative matrix \(\mathbf{X} \in \mathbb{R}^{M \times N}\) into two low-rank non-negative matrices, \(\mathbf{W} \in \mathbb{R}^{M \times k}\) and \(\mathbf{H} \in \mathbb{R}^{k \times N}\), such that \(\mathbf{X} \approx \mathbf{W} \mathbf{H}\), where $k$ is much smaller than $M $ and $N$. 

\textbf{NMF procedure.} One effective approach to find $\bf W$ and $\bf H$ is by utilizing Kullback-Leibler (KL) divergence as a discrepancy measure, which offers a sound statistical interpretation in applications involving counts or probabilities. The optimization aims to minimize the divergence between \(\mathbf{X}\) and its approximation \(\mathbf{W} \mathbf{H}\), which is given by~\cite{lee2001algorithms}:
\begin{align*}
    \mathbf{W}_{ij} \leftarrow \mathbf{W}_{ij} \frac{(\mathbf{X} \oslash (\mathbf{W} \mathbf{H})) \mathbf{H}^\top_{ij}}{{(\mathbf{1} \mathbf{H}^\top)}_{ij}} \\
    \mathbf{H}_{ij} \leftarrow \mathbf{H}_{ij} \frac{{\mathbf{W}^\top (\mathbf{X} \oslash (\mathbf{W} \mathbf{H}))}_{ij}}{(\mathbf{W}^\top \mathbf{1})_{ij}}
\end{align*}
where \(\oslash\) denotes element-wise division and \(\mathbf{1}\) denotes a one-matrix of \(\mathbf{X}\)'s size. These updates are applied iteratively, where each iteration improves the approximation of \(\mathbf{X}\) by reducing the KL divergence. The process is generally governed by a pre-determined number of iterations, or until a convergence criterion is reached. We use ${\bf W}, {\bf H} = {\bf NMF} ({\bf X}, {\bf W _{init}}, {\bf H_{init}}, T)$ to denote this iterative update procedure, where $T$ represents the number of updates.

\textbf{Robustness of NMF.} NMF is appreciated for its robustness in data reconstruction, particularly against noise. This robustness can be reflected via the triangle inequality~\cite{tversky1982similarity}:
\begin{align}
    \Vert \mathbf{X} + \delta - \mathbf{WH} \Vert \leq \Vert \mathbf{X} - \mathbf{WH} \Vert + \Vert \delta_2 \Vert \label{eq:triangle}
\end{align}
The inequality implies that any perturbation \(\delta\) to the data \(\mathbf{X}\) cannot induce a reconstruction error that exceeds the original reconstruction error by a margin of \( \| \delta\|\). 

It is more interesting to examine the robustness of the resulting $\bf W'$ and $\bf H'$ from a perturbed ${\bf X} + \delta$. Regarding that,  Laurberg's Theorem~\cite{Laurberg} provides a compelling mathematical foundation. Particularly, by denoting:
\begin{align*}
J_{(\mathbf{W}, \mathbf{H})}(\mathbf{W}', \mathbf{H}') := & \min_{\mathbf{D}, \mathbf{P}} \Vert \mathbf{W} - \mathbf{W}'(\mathbf{DP}) \Vert \\
&+ \Vert \mathbf{H} - (\mathbf{DP})^{-1}\mathbf{H}' \Vert
\end{align*}
where \(\mathbf{D}\) is a diagonal matrix and \(\mathbf{P}\) is a permutation matrix, we can restate the Laurberg's result as:

\textit{Assuming \(\mathbf{X} = \mathbf{WH}\) be a unique NMF. For any given \(\epsilon > 0\), there exists a \(\delta > 0\) such that for any nonnegative matrix \(\mathbf{Y} = \mathbf{X} + \mathbf{N}\) with \(\Vert \mathbf{N} \Vert < \delta\), we have $J_{(\mathbf{W}, \mathbf{H})}(\mathbf{W}', \mathbf{H}') < \epsilon$, where $[\mathbf{W}', \mathbf{H}'] = \argmin_{\mathbf{W}' \geq 0, \mathbf{H}' \geq 0} \Vert \mathbf{Y} - \mathbf{W}' \mathbf{H}' \Vert $.}

In other words, the Theorem shows that the perturbation's magnitude bounds the distortion in the factored matrices resulting from perturbed data, emphasizing the stability and robustness of NMF under near-optimal conditions.

\section{Latent Feature Attacks on NMF} \label{sect:method}
\begin{figure}
    \centering
    \includegraphics[width=0.99\linewidth]{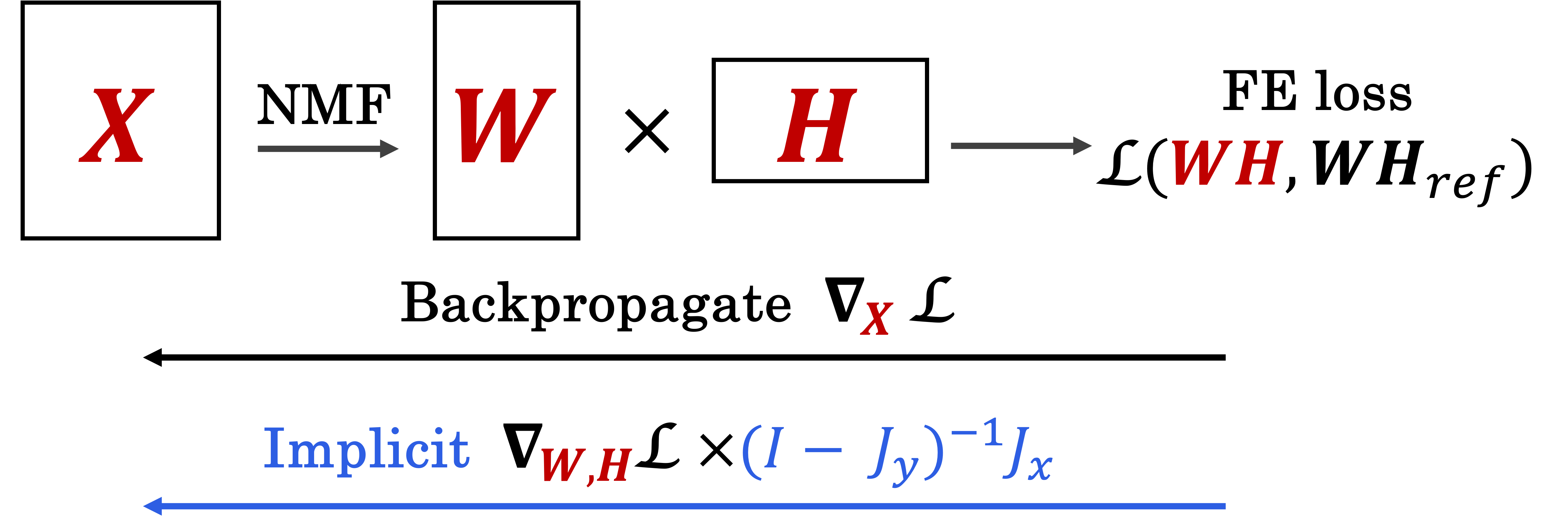}
    \caption{Adversarial gradient computation w.r.t FE loss (\ref{eq:FE}) via back-propagate and implicit methods. As the implicit does not need to backward the NMF, it results in the peak-memory advantage compared to the back-propagate.}
    \label{fig:overview}
\end{figure}

This section describes our proposed LaFA on NMF. Specifically, we introduce our proposed FE loss in Subsect.~\ref{subsect:FEloss}, and the gradient-ascent-based attacks in Subsect.~\ref{sec:AttAlg}. Fig.~\ref{fig:overview} provides an illustration of two proposed LaFA, called \textit{Back-propagation} and \textit{Implicit}. Both attacks compute the gradients of the FE loss w.r.t the input data $\bf X$ and utilize gradient-based methods to iteratively find the adversarial perturbation maximizing the FE. The {Back-propagation} method directly computes the gradient $\nabla_{\bf X} \mathcal{L}$ by reversing the NMF procedure, which demands extremely high peak-memory to store the gradient computational graph. In contrast, the {Implicit} method only need to backward to the feature matrices $\bf W$ and $\bf H$, significantly reducing the amount of memory required and enabling the scaling of attacks to scenarios with larger datasets. 

\subsection{Feature Error Loss} \label{subsect:FEloss}

We now elaborate on how to formulate a loss function capturing the feature errors between the NMF-extracted matrices $(\bf{W}_{NMF},\bf{H}_{NMF})$ and the true matrices $(\bf{W}_{true}, \bf{H}_{true})$ generating $\bf X$. We begin by denoting that error as a loss $\mathcal{L}$ taking two matrices  ${\bf WH_{NMF}}$ and ${\bf WH_{true}}$ as arguments: 
\begin{equation}
\text{FE} = \mathcal{L} \left({\bf WH_{NMF}},{\bf WH_{true}}\right) 
\label{eq:FE}
\end{equation}
Here, ${\bf WH} \in \mathbb{R}^{(M + N) \times k}$ is the concatenation of $\bf{W}$ and $\bf{H^{\top}}$ combined with a magnitude balancing operation: 
\begin{align*}
&{\bf WH}_i = {\bf concat}\left(\Bar{\bf{W}}_i,\Bar{\bf{H}}^\top_i\right) \quad \textup{where} \\
    \Bar{\bf{W}}_i &= {\bf{W}}_i\frac{\sqrt{\left\|{\bf{W}}_i\right\|\left\|{\bf H}_i\right\|}}{\left\|{\bf W}_i\right\|} \ \textup{,} \
    \Bar{\bf{H}}^\top_i = {\bf H}^\top_i\frac{\sqrt{\left\|{\bf W}_i\right\|\left\|{\bf H}_i\right\|}}{\left\|{\bf H}_i\right\|}
\end{align*}
with $i$ refers to the column of the matrices and $\left\|.\right\|$ is $L_2$ norm. For brevity, we denote the above construction of  $\bf{WH}$ from $\bf{W}$ and $\bf{H}$ by $\bf{WH} = \Bar{cat} (\bf{W}, \bf{H}) $. The magnitude balancing operation does not change the result of $\bf{W} \times \bf{H}$; however, it removes the ambiguity arising from the scaling of $\bf{W}$ and $\bf{H}$. Additionally, by including both $\bf{W}$ and $\bf{H}$ in the same norm operation, rather than in two separate terms as Laurberg's formulation~\cite{Laurberg}, the FE simplifies the features' alignment between $\bf{WH}_{true}$ and $\bf{WH}_{NMF}$. 

As identical $\bf{X}$s are recovered if the rows and columns of $\bf{W}$ and $\bf{H}$ are permuted in combination, a meaningful FE loss must minimize over all column-permutations of $\bf{WH}_{\bf NMF}$. To address this, a feature-wise error matrix $ {\bf FEM}$ are constructed as follow:
\begin{align*}
    {\bf FEM}\left(i,j\right) = \left\|{\bf WH_{NMF}}_i-{\bf WH_{true}}_j \right\|_F
\end{align*}
Then, the element-wise square of ${\bf FEM}$ can be fed into the Hungarian Algorithm~\cite{kuhn1955hungarian} to align $\bf{WH}_{NMF}$ so that $\left\|{\bf WH_{NMF}}-{\bf WH_{true}} \right\|$ is minimized. 

Consequently, we can express the FE loss $\mathcal{L}$ as:
\begin{align}
    & \text{FE} = \mathcal{L} \left({\bf WH_{NMF}},{\bf WH_{true}}\right) \\
    &= \min_{ \bf P}  \frac{\left\|\Bar{\bf cat} ( \bf P{\bf W_{NMF}}, \bf P {\bf H_{NMF}})-\bf{WH}_{true}\right\|_F}{\left\|\bf{WH}_{true}\right\|_F} 
\end{align}
where $\bf P$ is a permutation matrix.

Noting that squaring ${\bf FEM}$ linearizes the contribution of the difference of each element of $\bf{WH}$, thus allowing the linear sum assignment algorithm to correctly choose the minimum permutation without sampling all possible permutations. It would not be possible to utilize the Hungarian Algorithm to simplify the permutation problem if Laurberg's two-term definition of FE were used. 



\subsection{Latent Feature Attacks}
\label{sec:AttAlg}

With the FE loss (\ref{eq:FE}), the optimal direction for an adversarial feature attack can be computed. Given a ground-truth $\bf{WH}_{true}$, the FE can be considered as a function of $\bf{X}$, and the optimal adversarial direction is simply the gradient $\nabla_{\bf X} \mathcal{L}$. That gradient can be obtained by back-propagating the Multiplicative Updates of NMF (presented in subsection~\ref{sect:prelims}) and the FE loss. Then, the optimal distortion $ \varepsilon$ of a given magnitude producing the largest $\textup{FE}$ can be obtained via gradient-ascent algorithms.  This leads to the Back-propagation method, which will be described in Subsect.~\ref{subsect:bp}. However, this method demands a peak-memory usage proportional to the number of NMF iterations. This memory requirement is often infeasible for current computational capabilities, even for medium-sized datasets. As such, we propose another method, called {Implicit} Method utilizing the fixed-point condition of the NMF at convergence to implicitly compute the gradient (Subsect.~\ref{subsect:im}).  Since the memory requirement for the {Implicit} method is independent of the number of NMF iterations, it can scale the FE attack to larger datasets, effectively demonstrating the significant threat posed by NMF feature attacks.

\subsubsection{Back-propagate Feature Attack} \label{subsect:bp}

Algo.~\ref{algo:backprop_onestep} shows how to compute the adversarial direction for a feature attack using the Back-propagating method. That gradient then can be leveraged by Fast Gradient Signed Method (FGSM)~\cite{goodfellow2014explaining} or Projected Gradient Descent (PGD)~\cite{madry2017towards} to generate the adversarial $\bf \Tilde{X}$. However, the high memory requirement to backward the NMF iterations ${\bf W}, {\bf H} = {\bf NMF} ({\bf X}, {\bf W _{init}}, {\bf H_{init}}, T)$ (Line~\ref{algo:NMF}) hinders the practicality of the method. In fact, the gradients' computational graph for that step requires a peak-memory of $ T \times {O}(MN)$. Since the number of NMF's updates $T$ is typically $\approx 10^4$, it creates heavy burdens on computational resources and prevents the feasibility of the attack.



\begin{algorithm}[tb]
\caption{Adversarial gradients' computation on NMF via back-propagation}
\label{algo:backprop_onestep}
\textbf{Input:} \textbf{X}, $\bf{W}_{ref}$ $\bf{H}_{ref}$, and budget $\varepsilon$ \\
 \textbf{Parameters:} NMF iterations $T$ \\
 \textbf{Output:} Adversarial gradient $G$ 
 \begin{algorithmic}[1]
 \STATE $\bf{WH}_{\bf{ref}}=\Bar{cat} \left({\bf{W}}_{\bf{ref}} ,{\bf{H}}_{\bf{ref}}\right)$  \\
\STATE Randomly initialize $\bf{W}_{init}$ and $\bf{H}_{init}$
\STATE ${\bf W}, {\bf H} = {\bf NMF} ({\bf X}, {\bf W _{init}}, {\bf H_{init}}, T)$ \label{algo:NMF}
\STATE $\bf{WH} = \Bar{cat} (\bf{W}, \bf{H})$
\STATE $L = \mathcal{L} \left({\bf WH},{\bf WH_{ref}}\right)  $
\RETURN $G =  \nabla_{\bf X} L$  {\color{blue} \# Compute via backward to $\bf X$}
\label{algo:grad}
\end{algorithmic}
\end{algorithm}

\subsubsection{Implicit Method for Feature Attack}\label{subsect:im}

We now demonstrate our Implicit method to efficiently compute the gradient for feature attacks. The attack relies on the fixed-point condition of the NMF at convergence:
\begin{align}
   ({\bf W}, {\bf H}) = {\bf NMF} ({\bf X}, {\bf W}, {\bf H}, 1) \label{eq:nmf_update_one}
\end{align}
We denote $\bf x$ and $\bf y$ as the flattened vectors of $\bf X$, and $(\bf W , \bf H)$, respectively. We then can rewrite (\ref{eq:nmf_update_one}) as ${\bf \Tilde{f}}({\bf x}, {\bf y}) - {\bf y} = \bf 0$,
where ${\bf \Tilde{f}}$ is the NMF update with argument $\bf y$ instead of $\bf W$ and $\bf H$. By denoting $ \mathbf{f}(\mathbf{x}, \mathbf{y}) = \mathbf{\Tilde{f}}(\mathbf{x}, \mathbf{y}) - \mathbf{y} $, we have
\begin{align*}
     &\frac{d f_i}{d x_j} = \frac{\partial f_i}{\partial x_j} + \sum_k \frac{\partial f_i}{\partial y_k} \frac{d y_k}{d x_j} = 0 \\
     \Rightarrow & \frac{\partial f_i}{\partial x_j} + \sum_k \frac{\partial \Tilde{f}_i}{\partial y_k} \frac{d y_k}{d x_j} -
     \sum_k \frac{\partial y_i}{\partial y_k} \frac{d y_k}{d x_j}= 0 
 \end{align*}
 We can rewrite the above with the matrix's notations:
  \begin{align}
     & \frac{\partial \mathbf{f}}{\partial x_j} + J_y  \frac{d \mathbf{y}}{d x_j} - I  \frac{d \mathbf{y}}{d x_j} = 0 
     \Rightarrow \frac{\partial \mathbf{f}}{\partial x_j} + (J_y - I) \frac{d \mathbf{y}}{d x_j}  = 0 \nonumber \\
    & \Rightarrow \frac{d \mathbf{y}}{d \bf x} = -(J_y-I)^{-1} \frac{\partial \mathbf{f}}{\partial \bf x} = -(J_y-I)^{-1} J_x\label{eq:fixed}
\end{align}
where $J_y$ and $J_x$ are the Jacobian matrices with $ J_y[i,k] = \frac{\partial \Tilde{f}_i}{\partial y_k}$ and $J_x[i,k] = \frac{\partial \Tilde{f}_i}{\partial x_j}$. For (\ref{eq:fixed}), we use $ \frac{\partial {f}_i}{\partial x_j} = \frac{\partial \Tilde{f}_i}{\partial x_j} $.

We can see that (\ref{eq:fixed}) offers an alternative to compute the gradient. First, we use the Jacobians of one NMF update to compute the partial derivatives of $\bf W$ and $\bf H$ w.r.t. $\bf X$, i.e., $ J_y$, then multiply it with the gradient of the FE loss (\ref{eq:FE}) w.r.t.  $\bf W$ and $\bf H$ would give us the feature attack's gradient:
\begin{align*}
    G = - \nabla_{\bf W, \bf H} \mathcal{L} \times \left(\begin{bmatrix}
J_{\bf WW} & J_{\bf WH}\\ 
J_{\bf HW} & J_{\bf HH}
\end{bmatrix} - I \right)^{-1}  \times \begin{bmatrix}
J_{\bf WX} \\ 
J_{\bf HX} 
\end{bmatrix}
\end{align*}
This implicit computation scheme is summarized in Algo.~\ref{algo:implicit_onestep}. It can be seen that the peak-memory of the computation is for storing $J_x$, which is $O((M+N)DMN)$, and it is independent of the NMF's iterations $T$.

\begin{algorithm}[tb]
\caption{Adversarial gradients' computation on NMF via implicit function}
\label{algo:implicit_onestep}
\textbf{Input:} \textbf{X}, $\bf{W}_{ref}$ $\bf{H}_{ref}$, and budget $\varepsilon$ \\
 \textbf{Parameters:} NMF iterations $T$ \\
 \textbf{Output:} Adversarial gradient $G$ 
 \begin{algorithmic}[1]
 \STATE $\bf{WH}_{\bf{ref}}=\Bar{cat} \left({\bf{W}}_{\bf{ref}} ,{\bf{H}}_{\bf{ref}}\right)$  
\STATE Randomly initialize $\bf{W}_{init}$ and $\bf{H}_{init}$
\STATE ${\bf W}, {\bf H} = {\bf NMF} ({\bf X}, {\bf W _{init}}, {\bf H_{init}}, T)$ \\
 {\color{blue} \# Implicit gradients} \\
\STATE $J_{\bf WW}, J_{\bf WH}, J_{\bf WX}, J_{\bf HW}, J_{\bf HH}$ and $J_{\bf HX} \leftarrow$ \\ Jacobians of ${\bf NMF} ({\bf X}, {\bf W}, {\bf H}, 1)$
\STATE $J_y = \begin{bmatrix}
J_{\bf WW} & J_{\bf WH}\\ 
J_{\bf HW} & J_{\bf HH}
\end{bmatrix}$,  $J_x = \begin{bmatrix}
J_{\bf WX} \\ 
J_{\bf HX} 
\end{bmatrix}$
\STATE $G_{yx} = -(J_y-I)^{-1} J_x$
\STATE $L = \mathcal{L} \left({\bf WH},{\bf WH_{ref}}\right)  $
\STATE $G_{\mathcal{L}y} =  \nabla_{\bf W,H} L$  {\color{blue} \# Compute via backward to $\bf W, H$}
\RETURN $G = G_{\mathcal{L}y} \times  G_{yx} $
\end{algorithmic}
\end{algorithm}


\section{Illustrative Synthetic Experiments}\label{sect:example}



\begin{figure*}
\centering
\begin{subfigure}{.2\textwidth}
  \centering
  \includegraphics[height=1in]{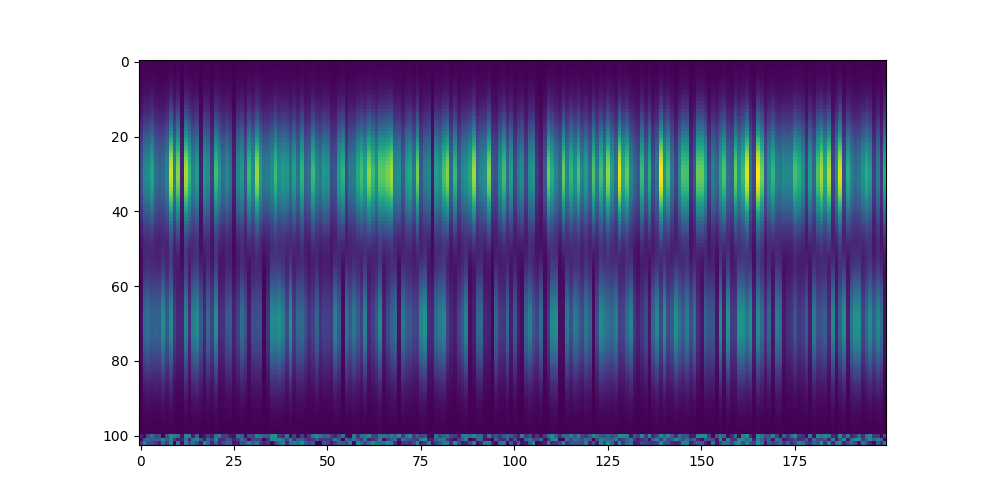}
  \centering
  \caption{Synthetic data $\bf X$.}
  \label{fig:X}
\end{subfigure}%
\begin{subfigure}{.39\textwidth}
  \centering
  \includegraphics[height=1in]{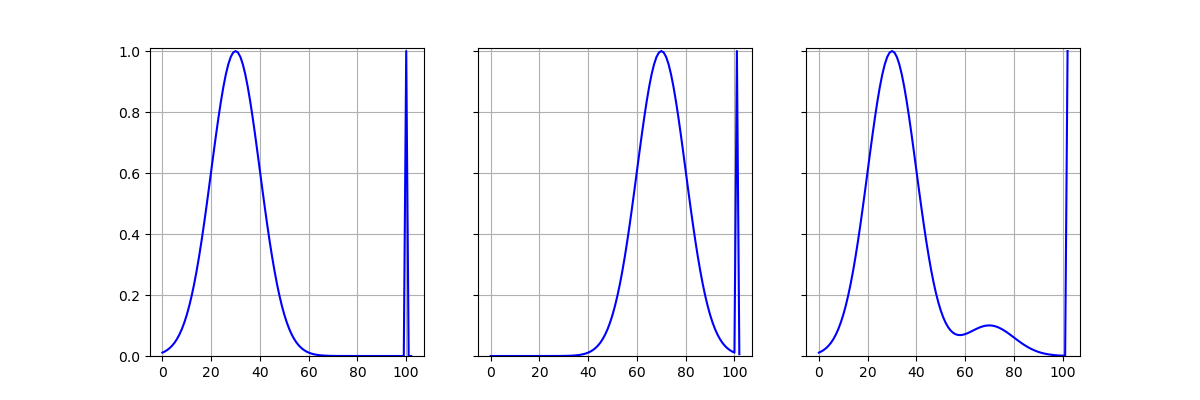}
  \centering
  \caption{Original $\bf W$.}
  \label{fig:Worig}
\end{subfigure}
\begin{subfigure}{.39\textwidth}
  \centering
  \includegraphics[height=1in]{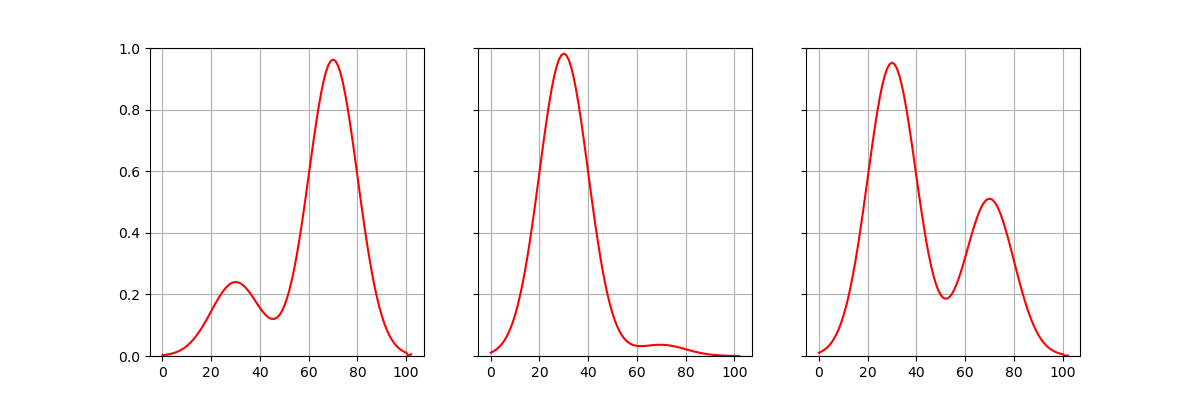}
  \caption{Reconstructed $\bf W$ of \textit{Removing-spike} adversarial.}
  \label{fig:Wpert}
\end{subfigure}

\begin{subfigure}{.49\textwidth}
  \centering
  \includegraphics[height=1.3in]{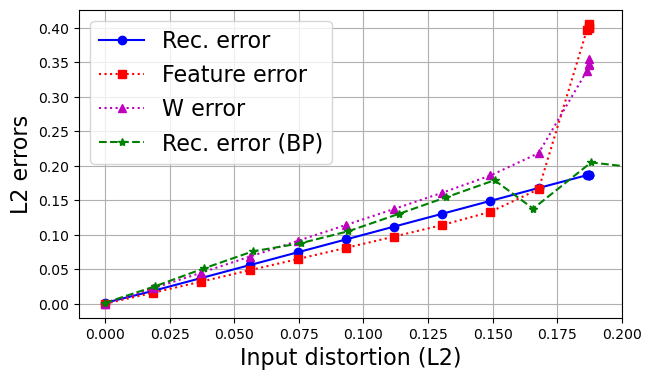}
  \caption{Distortions caused by \textit{Removing-spike} and \textit{Rec. loss} adversarial.}
  \label{fig:errors_rec}
\end{subfigure}
\begin{subfigure}{.49\textwidth}
  \centering
  \includegraphics[height=1.3in]{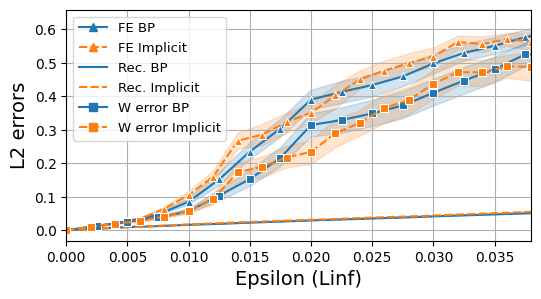}
  \centering
  \caption{FE and Reconstruction errors of proposed adversarial attacks.}
  \label{fig:errors_attack}
\end{subfigure}
\caption{Feature attacks on synthetic data of rank 3. While the reconstruction errors remain small, feature errors can be significantly large (Fig.~\ref{fig:errors_attack}). Notably, the Implicit method achieves a significant peak-memory advantage compared to Back-propagating (BP) method, i.e., 186.4Mbs compared to 278.2Mbs, while maintaining competitive attacking performance.}
\label{fig:orig_feat}
\end{figure*}

\begin{figure}
\centering
\includegraphics[width=.98\linewidth]{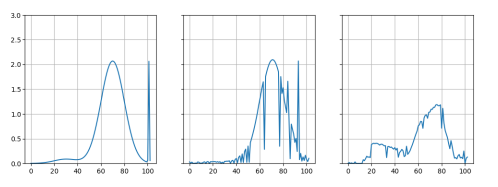}
 $\varepsilon = 0$ \quad \quad \quad \quad \quad $\varepsilon = 0.01$ \quad \quad \quad \quad $\varepsilon = 0.04$ 
\caption{The reconstruction of the first component of ${\bf W}$ in synthetic under PGD-$L_\infty$ Implicit attacks.}
\label{fig:syn_implicit}
\end{figure}

This section demonstrates the vulnerability of NMF against LaFA via a synthetic example. We consider a synthetic data ${\bf X} \in \mathbb{R}^{100 \times 200}$ (Fig.~\ref{fig:X}) with known ground-truth ${\bf W_{\bf true}} \in \mathbb{R}^{100 \times 3}$ (Fig.~\ref{fig:Worig}) and ${\bf H_{\bf true}} \in \mathbb{R}^{3 \times 200}$, and apply three attack/perturbation schemes on NMF when applied to $\bf X$. The first two columns of $\bf W$ are combinations of a Gaussian signal with a spike signal. The last column contains a linear combination of the first two Gaussian and an independent spike. Thus, $\bf W$ has rank 3. The matrix ${\bf H} \in$ generating $\bf X$ has its entries uniformly sampled from $[0,1)$. We now examine three following questions regarding NMF:

\textbf{Q1:} \textit{Find the direction of noise adding to $\bf X$ that induces high reconstruction error.}
As stated in Sect.~\ref{sect:prelims}, it is infeasible to inject a small perturbation to $\bf X$ that causes a large reconstruction error in NMF. Specifically, finding the direction maximizing reconstruction error corresponds to solve:
\begin{equation*}
\begin{aligned}
& \underset{||\epsilon||\leq \delta}{\text{maximize}}
& & \min_{\bf W \geq 0,H \geq 0} ||{\bf X} + \epsilon - {\bf W \times H}||_F.
\end{aligned}
\end{equation*}
Fig.~\ref{fig:errors_rec} shows our attempt to solve this with a gradient-ascent. The \textit{Rec. error (BP)} line shows the reconstruction error by using PGD directly backward on the reconstruction loss $||{\bf X} + \epsilon - {\bf W H}|| $. The results show that the reconstruction error is just slightly larger than the amount of noise injected on the input $\bf X$. This supports the theoretical analysis claiming that NMF is robust to reconstruction error.

\textbf{Q2:} \textit{Is there a noise direction that causes feature matrices to change the most and become unstable?} If the NMF decomposition of $\bf X$ is unique, then there is no \textit{bad} noise for arbitrarily small $\epsilon$~\cite{Laurberg}. However, when we are not dealing with an arbitrarily small epsilon, a \textit{bad} solution might exist.

To demonstrate that, we consider a perturbation $\bf \Tilde{X}$ of $\bf X$ such that its NMF's solutions would have high (or semantical) feature errors to those of $\bf X$. In particular, from $\bf W$ (Fig.~\ref{fig:Worig}) generating $\bf X$, we remove the spikes in the components of $\bf W$ and obtain a rank 2 ${\bf \Tilde{W}} \in \mathbb{R}^{100 \times 3}$. Then,  $\bf \Tilde{X}$ is set to ${\bf \Tilde{W}} \times \bf H$. $\bf \Tilde{X}$ is refereed as the \textit{Removing-spike} adversarial. Fig.~\ref{fig:Wpert} shows the resulting feature matrix when NMF is applied on $\bf \Tilde{X}$. It is significantly different from $\bf W$ not only in the absence of the spikes but also in the rank. 

To further study the \textit{Removing-spike}, we generate a set of perturbations along the direction from $\bf X$ to  $\bf \Tilde{X}$, i.e., $\{ \alpha {\bf \Tilde{X}} + (1 - \alpha) {\bf X} \}_{0\leq \alpha \leq 1}$, and compute the corresponding $\bf W$ and $\bf H$. The resulting reconstruction, FE, and the $L_2$ norm errorx of reconstructing $\bf W$ are plotted in Fig.~\ref{fig:errors_rec}. Interestingly, while the errors on features maintain proportional to the reconstruction errors when the input distortion is small, a significant spike in $\bf W$ errors and FE errors occur around $18\%$. This not only shows that the features' robustness proved by Laurberg for small noise does not hold for general noises, but also indicates our proposed FE loss (\ref{eq:FE}) has strong correlation to errors on reconstructed matrix $\bf W$.


\textbf{Q3:}  \textit{Find the noise that causes the highest feature errors}. The sharp increase of FE in Fig.~\ref{fig:errors_rec} suggests that small perturbation in a different direction may induce a much larger FE.  Specifically, the noise direction from $\bf X$ to  $\bf \Tilde{X}$ would require us to perturb about $18\%$ of the input to cause a sharp increase of $40\%$ in FE. The goal of our attacks is to find smaller noises that can induce larger feature errors, and, consequently, reveal the threat of feature attacks.

Fig.~\ref{fig:errors_attack} shows the performance of PGD-$L_\infty$ attacks~\cite{madry2017towards} leveraging our back-propagating (Algo.~\ref{algo:backprop_onestep}) and implicit gradients (Algo.~\ref{algo:implicit_onestep}) based on the FE loss. The results show that our attacks only require a perturbation of about $3\%$ of the input (at $\epsilon_\infty = 0.02$) to cause $40\%$ distortion in features. This importantly validates that NMF is vulnerable to feature attacks. Furthermore, this large distortion in features can not be detected solely from the reconstruction errors as the reconstruction errors remain approximately equal to the magnitude of the injected noise (as discussed in \textbf{Q1}).

Fig.~\ref{fig:syn_implicit} shows a more detailed look on the attacked features. While a small adversarial perturbation ($\epsilon = 0.01$) can distort the reconstructed $\bf W$ significantly, it does not change its rank (the spikes is preserved). The rank collapses to 2 and the spike disappears at $\epsilon= 0.04$. Thus, feature attacks not only create large distortions in terms of metric distances, but also alter the features' semantic.

\begin{figure}[ht]
\centering
\begin{subfigure}{.9\linewidth}
  \centering
  \includegraphics[width=.8\linewidth]{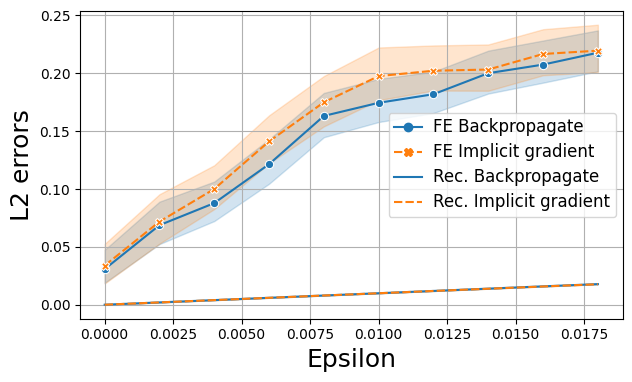}
  \caption{Feature and Reconstruction errors under $L_2$ attacks.}
  \label{fig:WTSI_L2}
\end{subfigure}%

\begin{subfigure}{.9\linewidth}
  \centering
  \includegraphics[width=.8\linewidth]{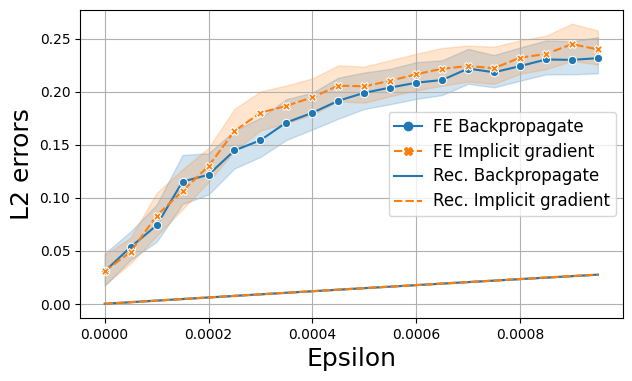}
  \caption{Feature and Reconstruction errors under $L_\infty$ attacks}
  \label{fig:WTSI_Linf}
\end{subfigure}%
 \vspace{-2mm}
\caption{Performance of FE attacks on WTSI dataset. Peak-memory BP: 109.5Mbs / Imp: 29.9Mbs.}
\label{fig:WTSI}
\end{figure}

\begin{figure}[ht]
\centering
\begin{subfigure}{.9\linewidth}
  \centering
  \includegraphics[width=.8\linewidth]{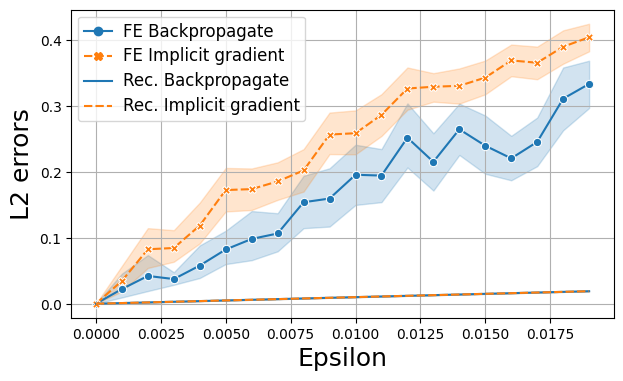}
  \caption{Feature and Reconstruction errors under $L_2$ attacks.}
  \label{fig:Face_L2}
\end{subfigure}%

\begin{subfigure}{.9\linewidth}
  \centering
  \includegraphics[width=.8\linewidth]{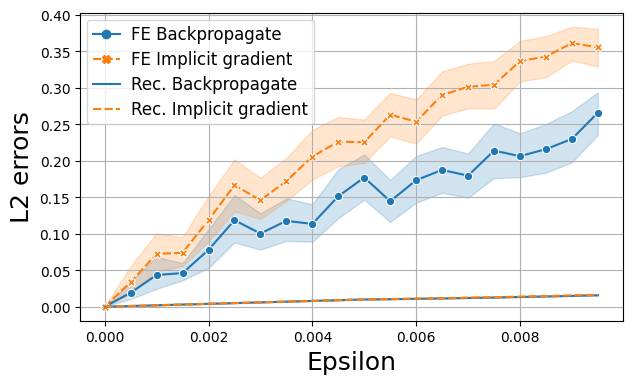}
  \caption{Feature and Reconstruction errors under $L_\infty$ attacks}
  \label{fig:Face_Linf}
\end{subfigure}%
 \vspace{-2mm}
\caption{Performance of FE attack on Face dataset. Peak-memory: BP: 8387.9Mbs / Imp: 6049.4Mbs.}
\label{fig:Face}
\end{figure}

\begin{figure}[ht]
\centering
  \centering
  \includegraphics[width=.7\linewidth]{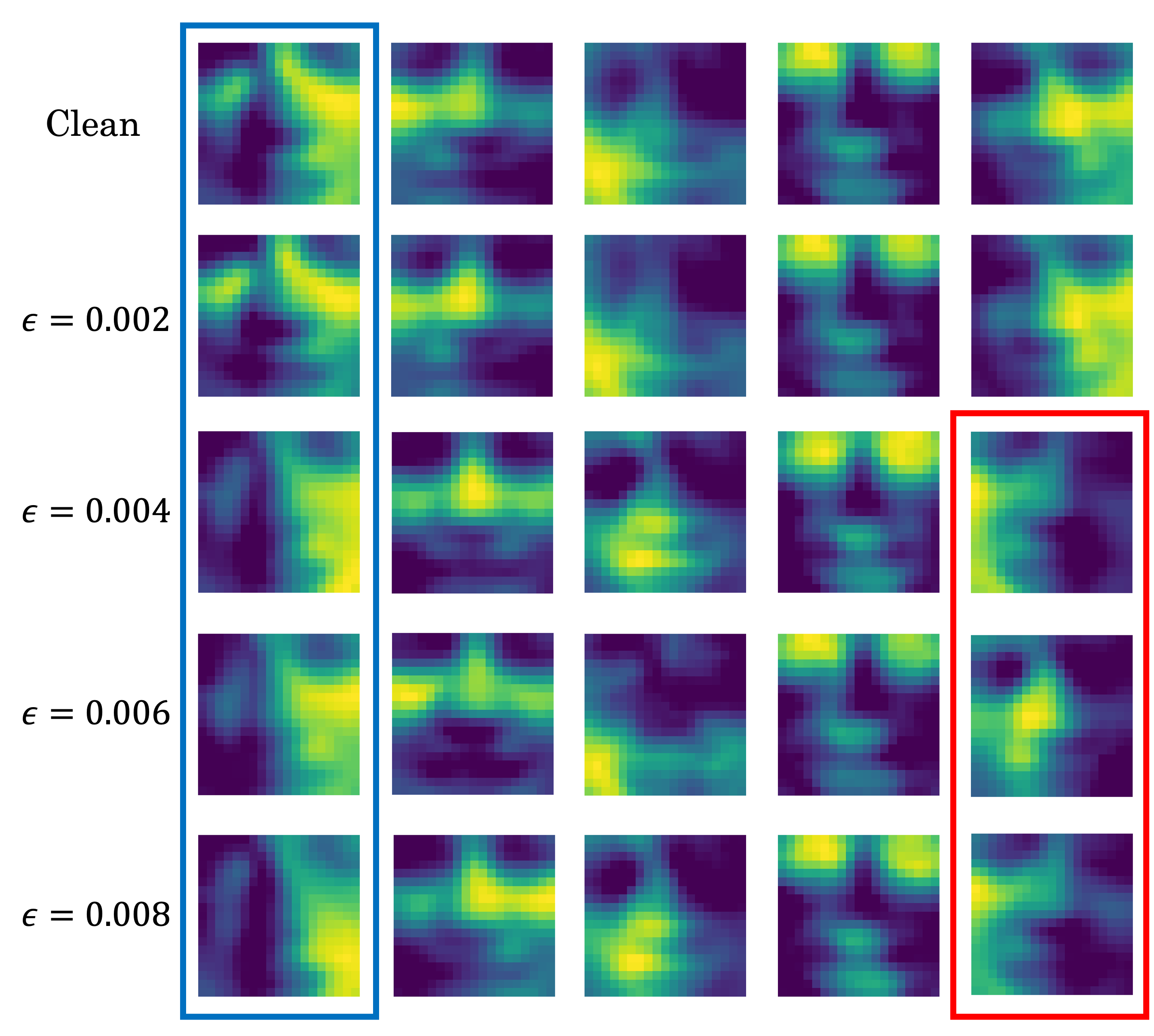}
   \vspace{-2mm}
\caption{Face's reconstructed $\bf W$ under $L_\infty$ FE attacks.}
\label{fig:Face_illu}
\end{figure}

\begin{figure}[ht]
\centering
\begin{subfigure}{.49\linewidth}
  \centering
  \includegraphics[width=.92\linewidth]{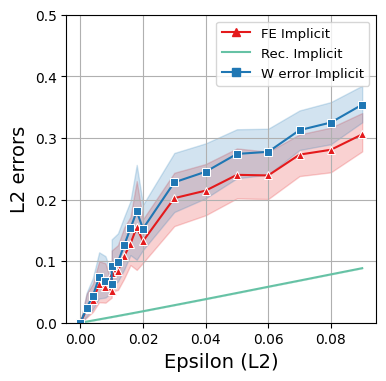}
  \caption{$L_2$ attack.}
  \label{fig:swim_L2}
\end{subfigure}%
\begin{subfigure}{.49\linewidth}
  \centering
  \includegraphics[width=.95\linewidth]{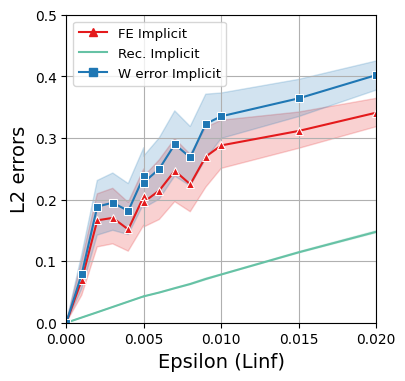}
  \caption{$L_\infty$ attack}
  \label{fig:swim_Linf}
\end{subfigure}%
\caption{Performance of FE attack on Swim dataset. Peak-memory approximately 3Gbs.}
\label{fig:swim}
\end{figure}

\begin{figure*}[ht]
\centering
  \centering
  \includegraphics[width=.99\linewidth]{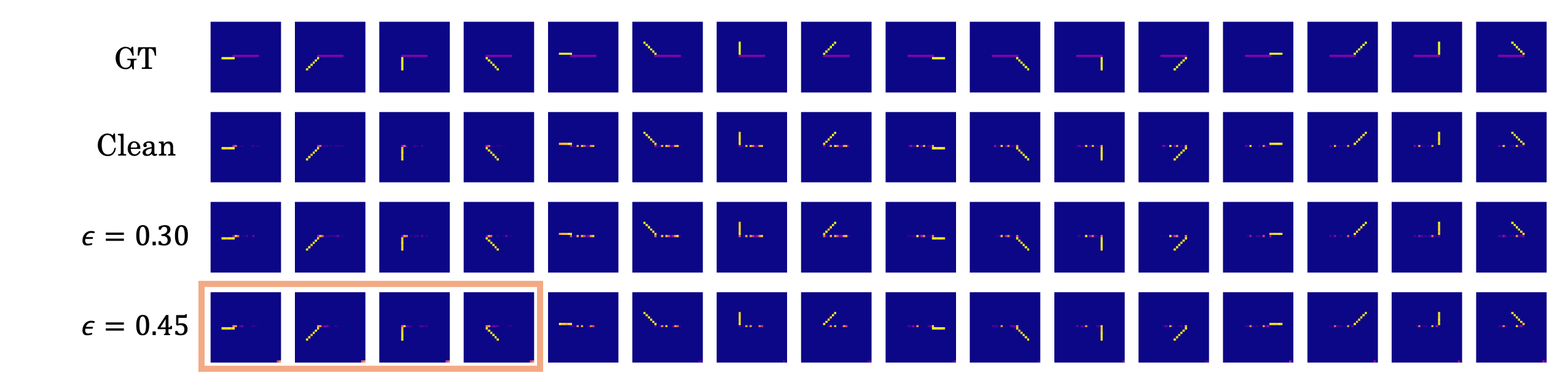}
\caption{Swim's reconstructed $\bf W$ under $L_2$ FE attacks.}
\label{fig:swim_illu}
\end{figure*}

\begin{figure}[ht]
\centering
\begin{subfigure}{.49\linewidth}
  \centering
  \includegraphics[width=.95\linewidth]{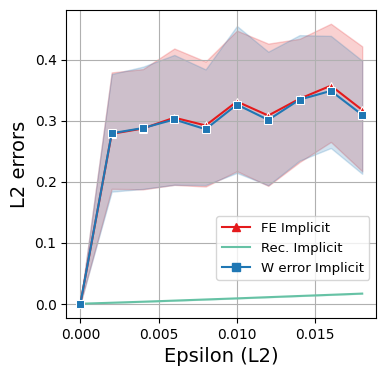}
  \caption{$L_2$ attack.}
  \label{fig:MNIST_L2}
\end{subfigure}%
\begin{subfigure}{.49\linewidth}
  \centering
  \includegraphics[width=.95\linewidth]{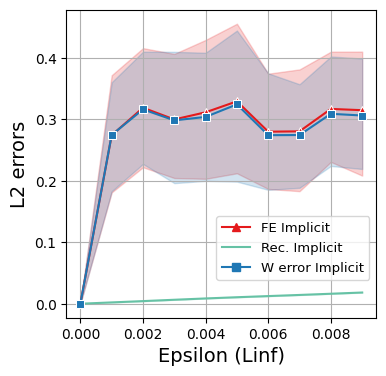}
  \caption{$L_\infty$ attack}
  \label{fig:MNIST_Linf}
\end{subfigure}%
 \vspace{-2mm}
\caption{Performance of FE attack on MNIST dataset. Peak-memory approximately 10Gbs.}
\label{fig:mnist}
\end{figure}
\begin{figure}[ht]
\centering
  \centering
  \includegraphics[width=.99\linewidth]{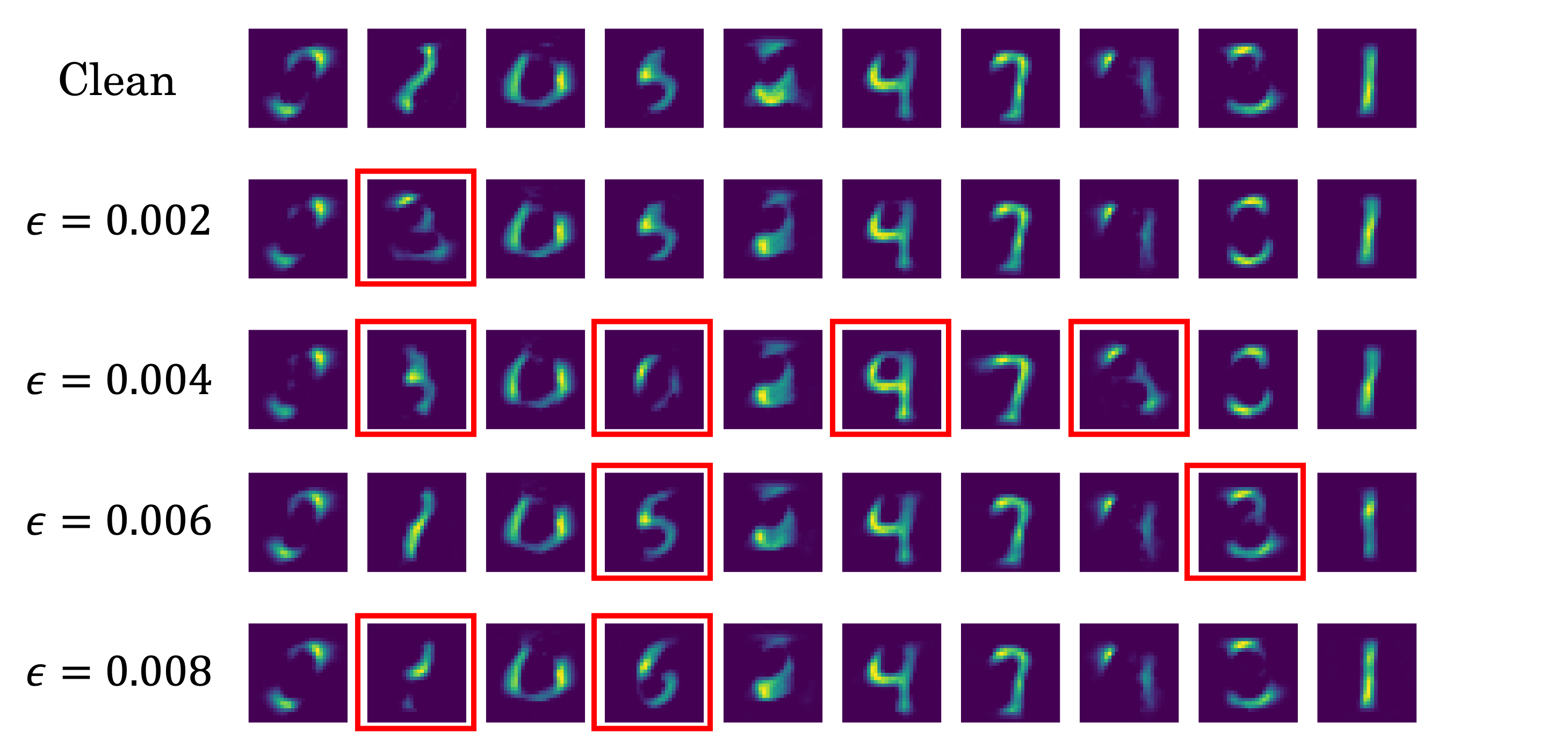}
  \vspace{-2mm}
\caption{MNIST's reconstructed $\bf W$ under  $L_\infty$ attacks.}
\label{fig:MNIST_illu}
\end{figure}

\section{Experimental Results} \label{sect:realworld}
This section demonstrates our findings on the vulnerabilities of NMF to LaFA in 4 real-world dataset: WTSI, Face, Swimmer and MNIST. The experiments consider the attacker has access to the data matrix $\bf X$ and its goal is to generate an adversarial noise resulting in high feature errors. All attacks utilize the PGD attack~\cite{madry2017towards} with $40$ steps leveraging on the gradients computed by either {Back-propagating} (Algo.~\ref{algo:backprop_onestep}) or {Implicit} (Algo.~\ref{algo:implicit_onestep}) methods. The entries of $\bf X$ are normalized between $0$ and $1$. 

The experiments are conducted on HPC clusters, equipped with AMD EPYC 7713 processors with 64 cores and 256 GB of RAM, and 4 NVIDIA Ampere A100 GPUs, each with 40 GB of VRAM.

\subsection{Results on WTSI dataset}
The WTSI~\cite{huang2018msignaturedb} is a genomic dataset featuring sequencing data from over 30 species, including a significant number of human genomic and cancer genome sequences. Experimental results on WTSI (Fig.~\ref{fig:WTSI}) reveal several key insights into the vulnerabilities of NMF to feature attacks. 
For both $L_2$ and $L_\infty$ attacks, the FE from {Back-propagation} and {Implicit} methods exhibit a linear increase in $L_2$ errors by increasing $\varepsilon$.
The {Back-propagation} consistently shows a slightly lower error trajectory compared to the {Implicit} method, suggesting it may be less effective at damaging the features. Notably, the {Back-propagation} requires a significantly higher peak-memory usage compared to the {Implicit} method, i.e., 109.5 MBs versus 29.9 MBs, highlighting the memory advantage of the {Implicit} method. These findings are crucial for assessing the vulnerability and designing protection strategies of genomic data to adversarial attacks.

\subsection{Results on Face Dataset}

The Face dataset~\cite{lee1999learning} comprises a set of 2,429 face and 4,548 non-face images. Due to the high computational complexity associated with back-propagating the NMF, we focus on 471 $19\times 19$-grayscale face images from the test set, forming ${\bf X} \in \mathbb{R}^{471 \times 361}$. We factorize $\bf X$ into 5 features. This choice is driven by the preliminary analysis indicating that this number of features captures the essential variability in the facial data while avoiding overfitting. 

NMF feature extraction vulnerability is evident in the Face dataset, as depicted in Fig.\ref{fig:Face_L2} and~\ref{fig:Face_Linf}. Notably, the {Implicit} method displays a much more significant advantage in attacking performance compared to Back-propagate compared to the results of WTSI. The visualization in Fig.~\ref{fig:Face_illu} showcases the effects of $L_{\infty}$ norm-based FE attacks on the reconstructed facial features matrix $\bf W$ from a specific dataset comprising face images. Each row represents different levels of perturbation's magnitude $\varepsilon$ in $L_{\infty}$, ranging from 0.002 to 0.008. The \textit{clean} row serves as a baseline, showing unperturbed latent features. As $\varepsilon$ increases, noticeable visual distortions appear in the reconstructed features, particularly highlighted in the blue box. These distortions indicate a degradation in feature integrity, affecting the clarity and structure of facial features. Moreover, the red box highlights the introduction of new feature components, underscoring a significant adversarial impact.

\subsection{Results on Swimmer Dataset}

The Swimmer dataset~\cite{donoho2003does} comprises 256 images depicting top-down representations of an individual swimming. This dataset is specifically designed to facilitate the exploration of sparse representations and the effectiveness of various signal-processing algorithms. We performed NMF factorization of the dataset into 16 features as suggested by~\cite{donoho2003does}.

The results on Swimmer is reported in Fig.~\ref{fig:swim}. Both $L_2$ and $L_\infty$ attacks show a significant distortions in $L_2$ errors as $\varepsilon$ increases. 
We cannot conduct {Back-propagation} attacks on the Swimmer due to memory constraint. On the other hand, the {Implicit} method demonstrates an escalation in error at a larger epsilon values compared to previous datasets. The reason is the  latent features $\bf W$ of Swimmer are much cleaner and more distinctive. Fig.~\ref{fig:swim_illu} exemplifies the degradation of reconstructed Swimmer's latent feature at higher noise. The sequence of images demonstrates the impact on the visual integrity on the recovered features: with large $\varepsilon$, the outlines and orientations of the swimmers become distorted and progressively less recognizable compared to the clean and ground truth images. This visual distortion is particularly significant at  $\varepsilon= 0.45$, at which some adversarial pixels begin to appear at the bottom corners of some features (orange box). This showcases the effectiveness of our adversarial attacks in disrupting the NMF’s ability to reconstruct the original latent features.

\subsection{Results on MNIST Dataset}

The MNIST dataset~\cite{lecun2010} is a collection of handwritten digits commonly used for training and testing image processing systems. It contains 70,000 $28 \times 28$-grayscale images of digits.
As the MNIST dataset comprises images across 10 classes, we factorized it into 10 features.

The results of Implicit PGD attack on MNIST in Fig.~\ref{fig:mnist} displays a sharp increase in feature errors under both $L_2$ and $L_\infty$ perturbations at a relatively low $\varepsilon$. 
The MNIST's extracted features (Fig.~\ref{fig:MNIST_illu}) under different $\varepsilon$ clearly illustrates the degradation of NMF. Starting from a baseline of clean, clear images, the introduction of even a small perturbation ($\varepsilon=0.002$) can affect the edges and finer details of the digits. As the perturbation grows, more pronounced visual artifacts appear, particularly distorting digits with complex structures such as '9', '3', and '5'. These digits start to merge with the background or deform significantly. 

\section{Conclusion and Future Work} \label{sect:conclusion}
\begin{table}[ht]
\caption{Running time of FE attacking methods.} \label{table:running_time}
 \resizebox{.99\linewidth}{!}{
\begin{tabular}{@{}cccccc@{}}
\toprule
\textbf{Dataset} & Synthetic & WTSI  & Face   & Swimmer & MNIST   \\ \midrule
Back-propagate    & 11.8s     & 35.9s & 76.4s  & N/A     & N/A     \\
Implicit         & 19.9s     & 17.7s & 297.5s & 396.0s    & 5214.3s
\end{tabular}
}
\end{table}

Throughout our investigation, we systematically explored the susceptibility of NMF to adversarial attacks across a spectrum of both synthetic and real-world datasets, with both {Back-propagation} and {Implicit} methods. Our findings demonstrated that adversarial perturbations could significantly impair the feature extraction capabilities of NMF, as evidenced by both norm-based metrics and direct visualizations of the corrupted features. The novel attack strategies introduced in this study also provide a significant step forward in understanding and enhancing the robustness of unsupervised learning frameworks. 

While our method has shown promising results in terms of performance and memory efficiency, the current implementation exhibits a running time complexity that may not be suitable for large-scale applications or real-time processing (Table~\ref{table:running_time}). To enhance the practicality and scalability of our approach, we aim to address the running time complexity of the proposed method in our future work.

\bibliographystyle{IEEEtran}
\bibliography{main}

\end{document}